\documentclass[10pt,twocolumn,letterpaper]{article}

\usepackage[pagenumbers]{cvpr}      

\usepackage{booktabs}       
\usepackage{amsfonts}       
\usepackage{nicefrac}       
\usepackage{microtype}      
\usepackage{xcolor}         
 
\usepackage{float}
\usepackage{tcolorbox}

\usepackage{latexsym}
\usepackage{graphicx}
\usepackage[para,online,flushleft]{threeparttable}
\usepackage{multicol}
\usepackage{multirow}
\usepackage{caption}
\usepackage{subcaption}
\usepackage{adjustbox}
\usepackage{colortbl}

\usepackage{wrapfig} 
\usepackage{inconsolata}
\usepackage{ulem}
\usepackage{tabularx}
\usepackage{amsmath}
\usepackage{amssymb}
\usepackage{longtable}
\usepackage{array}

\usepackage{makecell}
\usepackage{hhline}

\usepackage{amsthm}

\definecolor{cvprblue}{rgb}{0.21,0.49,0.74}
\usepackage[pagebackref,breaklinks,colorlinks,allcolors=cvprblue]{hyperref}

\def\confName{CVPR}
\def\confYear{2027}

\title{CAUSE: Context-Anchored State Editing for Modality-Consistent Multimodal Unlearning}

\author{
Xinghao Li$^{1}$, Chenchen Tan$^{1}$,
Youyang Qu$^{2,3}$, Shujie Cui$^{1}$,\\
Xingliang Yuan$^{4}$, Cunjian Chen$^{5}$,
Chi Liu$^{6}$, Longxiang Gao$^{2,3}$\\[1em]
$^{1}$Faculty of Information Technology, Monash University, Australia\\
$^{2}$Qilu University of Technology (Shandong Academy of Sciences), China\\
$^{3}$Shandong Computer Science Center
(National Supercomputer Center in Jinan), China\\
$^{4}$School of Computing and Information Systems,
The University of Melbourne, Australia\\
$^{5}$Department of Data Science and AI, Monash University, Australia\\
$^{6}$Faculty of Data Science, City University of Macau, Macao SAR, China
}

\begin{document}
\maketitle
\begin{abstract}
Multimodal large language models (MLLMs) can disclose memorized factual associations through both visual evidence and textual query information. Achieving modality-consistent forgetting remains a fundamental challenge in multimodal unlearning, as suppressing target knowledge through one modality does not necessarily prevent its recovery through another. Moreover, the entanglement of target and retained knowledge in shared representations complicates selective forgetting. We propose \textbf{CAUSE}, \underline{C}ontext-\underline{A}nchored \underline{U}nlearning via \underline{S}tate \underline{E}diting, a multimodal unlearning framework that establishes the context-to-answer transition as a shared intervention interface across modalities. CAUSE calibrates factual suppression according to identity-related differences in predictive support and adaptively coordinates modality-specific suppression objectives. To preserve non-target knowledge, CAUSE combines multi-view distillation with modality-specific gradient protection while keeping the pretrained backbone frozen. Experiments on MLLMU-Bench and UMU-Bench demonstrate that CAUSE can achieve effective factual suppression across multiple tasks, including classification, cloze completion, and generation, with more consistent forgetting across visual and textual access routes. It also achieves a stronger unlearning-retention trade-off than representative LLM and MLLM unlearning baselines.
\end{abstract}

\section{Introduction}
\label{sec:intro}

Large language models (LLMs)~\cite{brown2020language,chowdhery2023palm} and multimodal large language models (MLLMs)~\cite{liu2024improved,yin2025rod} have enabled increasingly capable generation and reasoning over linguistic and visual information~\cite{ouyang2022training,li2023blip}. However, their ability to memorize factual associations from large-scale training data raises concerns about sensitive information disclosure, intellectual property, and data governance~\cite{kandpal2022deduplicating,ICLR2025_cce0e917}. In MLLMs, information associated with an individual may be recovered through a textual identity reference, an image depicting that individual, or a combination of visual and linguistic evidence~\cite{tomekcce2024private,chen2025unveiling}. Such information disclosure motivates machine unlearning, which aims to suppress designated knowledge while preserving the model's competence on retained information~\cite{fan2024salun,yuancloser}.

While machine unlearning has been extensively studied in LLMs, extending it to MLLMs introduces additional challenges because the same knowledge can be accessed through different input modalities. Conventional methods, such as gradient ascent (GA)~\cite{jang-etal-2023-knowledge}, NPO~\cite{zhang2024negative}, and SimNPO~\cite{NEURIPS2025_02443e4e}, primarily formulate unlearning objectives for text-conditioned language models. MLLM-specific approaches further address visual and cross-modal knowledge: SIU~\cite{li2024single} employs single-image fine-tuning to suppress visual concept recognition, whereas MANU~\cite{liu2025modality} uses modality-aware neuron pruning for knowledge unlearning. Nevertheless, UMU-Bench~\cite{wang2025umu} reveals inconsistencies in forgetting effectiveness between multimodal and text-only inputs and investigates joint unlearning across both modalities. CMJU~\cite{yang2026cmju} further improves forgetting consistency through modality-aware parameter selection and coordinated gradient updates. Despite these advances, a fundamental question remains: if multimodal and text-only queries access the same factual knowledge, why does forgetting induced through one modality not necessarily transfer to the other?

\begin{figure}[t]
    \includegraphics[width=0.45\textwidth]{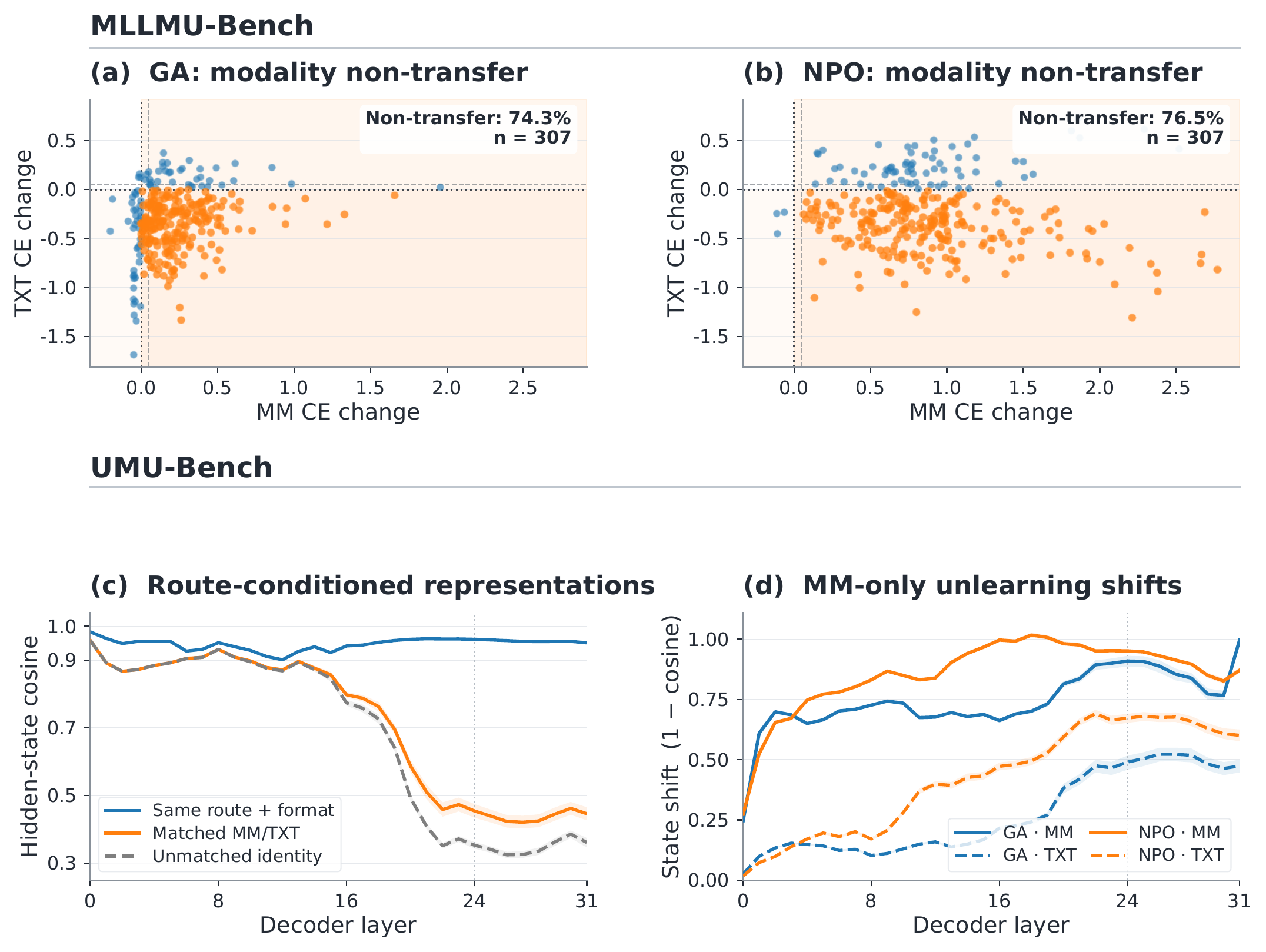}
    \caption{\label{fig:scatter} \small (a,b) Changes in target-answer cross-entropy (CE) following multimodal-only unlearning with GA and NPO on MLLMU-Bench. Non-transfer denotes the fraction of all paired samples exhibiting increased CE under multimodal (MM) queries ($\Delta\mathrm{CE}^{\mathrm{mm}}>0$) but unchanged or decreased CE under text-only (TXT) queries ($\Delta\mathrm{CE}^{\mathrm{txt}}\leq0$). (c) Layer-wise cosine similarity of contextual representations under matched route and query format, matched-identity MM/TXT, and unmatched-identity conditions on UMU-Bench. (d) Layer-wise representation shifts after multimodal-only unlearning on UMU-Bench.}
\end{figure}

To investigate this question, we conduct behavioral and representation-level analyses using MLLMU-Bench~\cite{liu2025protecting} and UMU-Bench~\cite{wang2025umu}, as illustrated in Fig.~\ref{fig:scatter}. At the behavioral level, we observe that conventional unlearning methods trained on multimodal data exhibit limited forgetting transfer to text-only inputs. More importantly, our representation-level analysis reveals that multimodal and text-only queries referring to the same factual knowledge induce distinct contextual representations and undergo asymmetric representational changes during multimodal-only unlearning. These findings suggest that semantic equivalence does not necessarily imply representational alignment or consistent forgetting, motivating explicit suppression objectives across both input modalities.

However, joint unlearning across modalities \cite{wang2025umu} alone does not guarantee sufficient or selective forgetting. Uniformly aggregating modality-specific objectives may obscure differences in their optimization progress, while aggressive parameter updates may disrupt retained predictions~\cite{NEURIPS2025_8e8388ec,shi2025muse}. Despite their representational differences, multimodal and text-only inputs share a common autoregressive prediction mechanism. The final context-position hidden state contributes to the first answer-token prediction, while subsequent states support successive predictions. This shared computational structure motivates us to anchor state interventions at the context-to-answer transition and extend them along the answer prediction process, without requiring explicit alignment of modality-specific representations or direct modifications to preceding context states.

Motivated by these findings, we propose \textbf{CAUSE} (\underline{C}ontext-\underline{A}nchored \underline{U}nlearning via \underline{S}tate \underline{E}diting), an unlearning framework that couples shared state editing with modality-adaptive factual suppression to address modality-inconsistent forgetting. Specifically, CAUSE anchors shared low-rank state corrections at the context-to-answer transition and extends them along the autoregressive answer prediction process. To accommodate fact- and modality-specific forgetting requirements, CAUSE calibrates suppression targets using identity-related differences in the original model's predictive support and regulates both the absolute support and relative confidence of target facts. The resulting modality-specific forgetting objectives are adaptively coordinated to promote consistent factual suppression across both modalities. To preserve non-target knowledge, CAUSE further integrates multi-view distillation with retention-aware gradient projection while keeping the pretrained backbone frozen. We evaluate CAUSE on MLLMU-Bench~\cite{liu2025protecting} and UMU-Bench~\cite{wang2025umu} across multiple tasks, including classification, cloze completion, and generation. The experimental results demonstrate that CAUSE achieves more consistent forgetting across modalities while maintaining a favorable balance between strong factual suppression and knowledge retention.
\section{Related Work}
\label{sec:related}

\paragraph{LLM Unlearning.}
Machine unlearning in LLMs aims to suppress designated knowledge while preserving model utility. Early approaches employ gradient ascent (GA) to reduce the likelihood of target information~\cite{jang-etal-2023-knowledge}, while KL-based objectives~\cite{nguyen2020variational} constrain deviations from the original model. NPO~\cite{zhang2024negative} and SimNPO~\cite{NEURIPS2025_02443e4e} formulate unlearning through negative preference optimization. Beyond output-level optimization, RMU~\cite{wmdp} redirects target representations towards random directions while preserving retained representations. However, these methods are primarily designed for text-conditioned models and do not explicitly address modality-dependent knowledge access.

\begin{figure*}[!t]
    \includegraphics[width=0.85\textwidth]{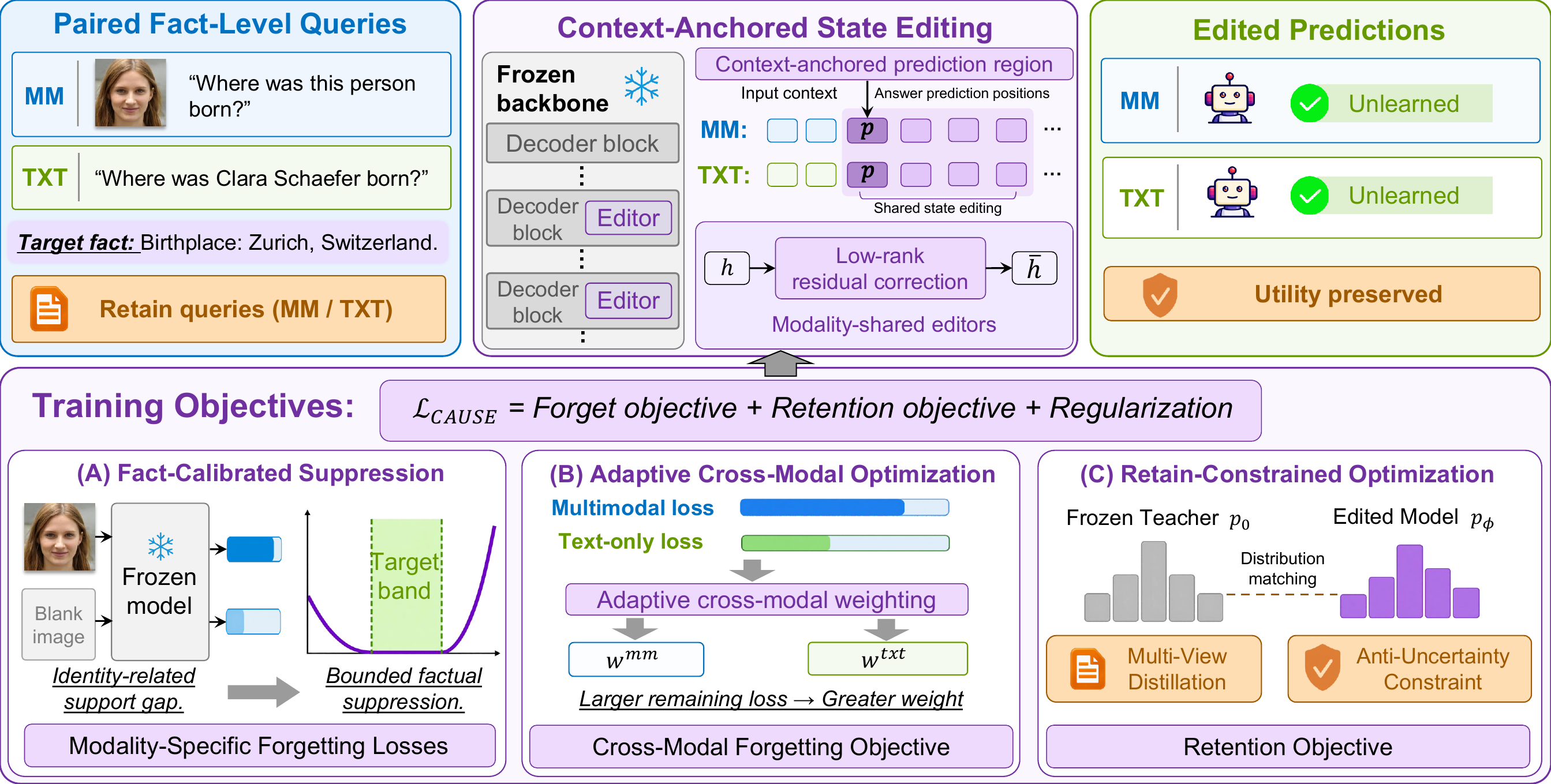}
    \centering
    \caption{\label{fig:method} \small Overview of the proposed CAUSE framework. CAUSE applies modality-shared low-rank state corrections from the context boundary along the answer prediction process while keeping the pretrained backbone frozen. (A) Fact-calibrated suppression regulates target factual support. (B) Adaptive cross-modal optimization coordinates modality-specific suppression objectives by emphasizing the larger remaining loss. (C) Retain-constrained optimization preserves non-target predictions through multi-view distillation, anti-uncertainty supervision, and retention-aware gradient projection. The modality-specific forgetting losses in (A) are adaptively aggregated in (B) and combined with the retention objective in (C) to optimize the shared state editors.}
\end{figure*}

\paragraph{MLLM Unlearning.}
MLLM unlearning addresses knowledge associated with both visual and textual information. MultiDelete~\cite{cheng2024multidelete}, MMUnlearner~\cite{huo2025mmunlearner}, and SIU~\cite{li2024single} suppress targeted knowledge through multimodal association removal or fine-tuning. Beyond fine-tuning, MANU~\cite{liu2025modality} and MIP-Editor~\cite{li2026crossmodal} localize knowledge-related neurons and pathways for selective intervention, while EMMU~\cite{zhang2025emmu} investigates information-level unlearning. MLLMU-Bench~\cite{liu2025protecting} and CLEAR~\cite{dontsov2025clear} further facilitate the evaluation of multimodal forgetting and retention. However, these works largely overlook a critical challenge in MLLM unlearning: suppressing target knowledge under multimodal inputs does not necessarily prevent its recovery through text-only queries.

\paragraph{Modality-Consistent Unlearning.}
UMU-Bench~\cite{wang2025umu} reveals modality-inconsistent forgetting and investigates joint multimodal and text-only unlearning. CMJU~\cite{yang2026cmju} further addresses this inconsistency through modality-aware parameter selection and gradient coordination. However, existing cross-modal unlearning approaches still face a fundamental trade-off between forgetting effectiveness and knowledge retention. Insufficient suppression may leave target facts recoverable through alternative modalities, whereas aggressive updates to shared model parameters can substantially impair retained knowledge. This highlights the need to regulate not only the coverage of forgetting across modalities but also the extent of factual suppression under each input condition. To this end, CAUSE integrates context-anchored state editing with fact-calibrated and modality-adaptive suppression, enabling targeted forgetting while limiting unintended changes to retained predictions.
\section{Method}
\label{sec:method}

\subsection{Overview and Problem Formulation}
\label{sec:cause_formulation}

As illustrated in Fig.~\ref{fig:method}, CAUSE establishes the context-to-answer transition as a shared intervention interface across input modalities. It applies low-rank state corrections along the autoregressive answer prediction process and optimizes the shared editors through modality-adaptive factual suppression and retain-constrained optimization, while keeping the pretrained backbone frozen.

Let $\mathcal D_F$ denote the set of target facts to be forgotten, where each fact is represented as $z_i=(e_i,a_i,v_i)$, with entity $e_i$, attribute $a_i$, and associated value $v_i$. We consider two query conditions for each fact:
\begin{equation}
\label{eq:cause_inputs}
x_i^{\mathrm{mm}}=(I_i,q_i^{\mathrm{ref}}),\qquad
x_i^{\mathrm{txt}}=q_i^{\mathrm{name}}.
\end{equation}
Here $I_i$ is an image depicting entity $e_i$, $q_i^{\mathrm{ref}}$ refers to the entity in the image, and $q_i^{\mathrm{name}}$ explicitly identifies the same entity by name. Both queries request the same attribute value $v_i$. We use $r\in\{\mathrm{mm},\mathrm{txt}\}$ to index the two input conditions and denote a query under formulation $t$ by $x_{i,t}^r$. Let $f_{\theta_0}$ be the original MLLM with pretrained parameters $\theta_0$. CAUSE keeps $\theta_0$ fixed and learns shared intervention parameters $\phi$, yielding the edited model $f_{\theta_0,\phi}$. These parameters are jointly optimized for factual suppression on $\mathcal D_F$ and knowledge preservation on an equally sized retain dataset $\mathcal D_R$.

\subsection{Context-Anchored State Editing}
\label{sec:cause_edit}

In autoregressive MLLMs, answer prediction states incorporate modality-conditioned contextual information through causal attention. The final context-position hidden state contributes to the first answer-token prediction, while subsequent hidden states support successive predictions. Recent studies suggest that contextual representations contain information predictive of subsequent responses~\cite{dong2025emergent,tan2026geometric}. Accordingly, CAUSE anchors state corrections at the context boundary and extends them along the answer prediction process, restricting direct interventions to prediction-relevant states while leaving preceding context states unedited.

For a given input $x_{i,t}^r$, let $N$ denote the decoder sequence length after multimodal token expansion and $m$ the number of appended answer tokens. Under zero-based indexing, the context boundary is $p=N-m-1$, and the intervention positions are $\mathcal P=\{p,\ldots,p+m-1\}$. During generation, the boundary remains fixed, and $\mathcal P$ extends along the generated answer prefix to include the current next-token prediction position.

Let $h_j^{(\ell)}\in\mathbb R^d$ denote the output hidden state at position $j$ in decoder layer $\ell$. CAUSE applies a low-rank residual correction at selected decoder layers $\ell\in\mathcal S$:
\begin{equation}
\label{eq:cause_edit}
\bar h_j^{(\ell)}
=
h_j^{(\ell)}
+
\mathbb I[j\in\mathcal P]
\frac{\alpha}{k}
\big(h_j^{(\ell)}A_\ell\big)B_\ell,
\end{equation}
where $A_\ell\in\mathbb R^{d\times k}$ and $B_\ell\in\mathbb R^{k\times d}$ are trainable matrices, $d$ is the hidden dimension, $k$ is the intervention rank, and $\alpha$ controls the residual scale. The trainable parameters are $\phi=\{A_\ell,B_\ell\}_{\ell\in\mathcal S}$, with $B_\ell$ initialized to zero.

The correction matrices are shared across modalities, while the resulting corrections depend on modality-conditioned hidden states. This design operates on a common prediction interface without requiring alignment of modality-specific representations. Restricting direct interventions to answer prediction positions further limits unnecessary modifications to preceding context states.

\subsection{Fact-Calibrated Suppression Across Modalities}
\label{sec:cause_forget}

To guide the shared state corrections, CAUSE calibrates the required reduction in factual support using identity-related differences in the original model's predictive support and adaptively coordinates the resulting modality-specific suppression objectives.

\paragraph{Identity-calibrated suppression.}

Given a target fact $z_i$, let $\mathcal U_i$ denote its training-side answer expressions. For $u\in\mathcal U_i$, let $y_{i,u}$ be its tokenized sequence and $\mathcal T_{i,u}$ the positions of its fact-bearing tokens. We define the factual support score as
\begin{equation}
\label{eq:cause_score}
s_\phi(x_i^r,u)
=
\frac{1}{|\mathcal T_{i,u}|}
\sum_{j\in\mathcal T_{i,u}}
\log p_{\theta_0,\phi}
\big(y_{i,u,j}\mid x_i^r,y_{i,u,<j}\big),
\end{equation}
where $p_{\theta_0,\phi}$ denotes the edited model's conditional token probability. The sensitivity of factual predictions to entity identification may vary across facts and input modalities. To estimate this variation, CAUSE constructs an identity-reduced reference $\widetilde x_i^r$ by reducing the identifying visual or textual cues while preserving the queried attribute. The identity-related support gap is
\begin{equation}
\label{eq:cause_gap}
g_{i,r,u}
=
s_0(x_i^r,u)-s_0(\widetilde x_i^r,u),
\end{equation}
where $s_0$ denotes the factual support assigned by the original model. The resulting gap serves as a fact- and modality-specific calibration signal reflecting changes in predictive support under reduced identity information.

For expressions satisfying $s_0(x_i^r,u)\geq s_{\min}$, we define the desired reduction and suppression target as
\begin{equation}
\label{eq:cause_target}
\begin{aligned}
d_{i,r,u}
&=
\begin{cases}
\min\{g_{i,r,u},d_{\max}\},
& g_{i,r,u}\geq\eta,\\
\min\{d_{\mathrm{fb}},d_{\max}\},
& g_{i,r,u}<\eta,\ u\in\mathcal U_i^{\mathrm{core}},
\end{cases}\\
b_{i,r,u}
&=s_0(x_i^r,u)-d_{i,r,u}.
\end{aligned}
\end{equation}
Here $s_{\min}$ and $\eta$ specify the minimum initial support and identity-related gap, respectively, while $d_{\max}$ caps the desired reduction. For eligible expressions with $\eta\leq g_{i,r,u}\leq d_{\max}$, the calibrated target reduces to $b_{i,r,u}=s_0(\widetilde x_i^r,u)$, adapting the suppression strength to the observed variation in factual support. When $g_{i,r,u}<\eta$, the observed support gap may provide an insufficient calibration signal. To maintain suppression of expressions that directly convey the target factual value, a predefined reduction $d_{\mathrm{fb}}$ is applied to eligible core expressions in $\mathcal U_i^{\mathrm{core}}$, subject to the same upper bound $d_{\max}$. Expressions that do not satisfy the eligibility conditions are excluded from calibrated suppression.

To control the extent of prediction changes, CAUSE penalizes deviations from a calibrated support interval:
\begin{equation}
\label{eq:cause_bounded}
\begin{aligned}
\ell_{i,r,u}
={}&[s_\phi(x_i^r,u)-b_{i,r,u}]_+^2\\
&+\gamma[b_{i,r,u}-\delta-s_\phi(x_i^r,u)]_+^2,
\end{aligned}
\end{equation}
where $[a]_+=\max(a,0)$, $\delta>0$ defines the permitted margin below the target, and $\gamma>0$ controls the lower-bound penalty. The two terms penalize insufficient and excessive suppression, respectively, yielding a zero-loss interval $[b_{i,r,u}-\delta,b_{i,r,u}]$. To prioritize insufficiently suppressed factual content, we retain eligible core expressions and adaptively select additional eligible expressions with the largest residual support $s_\phi(x_i^r,u)-b_{i,r,u}$. Let $\mathcal U_{i,r}^{\star}$ denote the resulting set. The calibrated suppression objective is $\mathcal L_{\mathrm B}^{i,r}=|\mathcal U_{i,r}^{\star}|^{-1}\sum_{u\in\mathcal U_{i,r}^{\star}}\ell_{i,r,u}$, with $\mathcal L_{\mathrm B}^{i,r}=0$ when the set is empty.

\paragraph{Factual confidence and response behavior.}

Calibrated support suppression establishes fact- and modality-specific targets for absolute support reduction. Complementing this calibration, CAUSE regulates the relative predictive preference for target facts and encourages uncertainty rather than unsupported factual substitutions.

For an answer sequence $y$, its sequence-level support is $\bar s_\phi(x,y)=|y|^{-1}\sum_{j=1}^{|y|}\log p_{\theta_0,\phi}(y_j\mid x,y_{<j})$. Given a candidate set $\mathcal C_{i,t}^r=\{y_0,\ldots,y_{K-1}\}$ containing the target answer $y_0$ and alternative values of the same attribute drawn from training data, we define the relative factual confidence as
\begin{equation}
\label{eq:cause_confidence}
\pi_{i,r,t}
=
\frac{\exp(c_0/T)}
{\exp(c_0/T)+
\sum_{j=1}^{K-1}\exp(\operatorname{sg}(c_j)/T)},
\end{equation}
where $c_j=\bar s_\phi(x_{i,t}^r,y_j)$, $T>0$ is a temperature parameter, and $\operatorname{sg}(\cdot)$ denotes stop-gradient. Detaching alternative candidate scores allows confidence suppression without directly reinforcing incorrect factual values. We penalize excessive target confidence by
\begin{equation}
\label{eq:cause_deconfidence}
\mathcal L_{\mathrm C}^{i,r,t}
=
\frac{1}{2}
\left[\pi_{i,r,t}-\left(\frac{1}{K}+\epsilon\right)\right]_+^2,
\end{equation}
where $\epsilon$ specifies the permitted margin above the uniform confidence level $1/K$. CAUSE further encourages an uncertainty response $y_\bot$ through $\mathcal L_{\mathrm U}^{i,r,t}=-\bar s_\phi(x_{i,t}^r,y_\bot)$.

\paragraph{Adaptive cross-modal suppression.}

The preceding objectives define the desired changes in factual predictions under each input condition. We combine them into a modality-specific forgetting objective:
\begin{equation}
\label{eq:cause_route_total}
\mathcal L_i^r
=
\mathcal L_{\mathrm B}^{i,r}
+
\sum_{t\in\mathcal Q_i^r}
\left(
\lambda_t\mathcal L_{\mathrm U}^{i,r,t}
+
\mu_t\mathcal L_{\mathrm C}^{i,r,t}
\right),
\end{equation}
where $\mathcal Q_i^r$ denotes the query formulations for supervision, and $\lambda_t$ and $\mu_t$ weight uncertainty supervision and confidence suppression, respectively. The confidence term is omitted when candidate-based supervision is not applied. To accommodate differences in optimization progress across modalities, CAUSE adopts a smooth worst-case objective that adaptively emphasizes the modality with a larger remaining loss while maintaining supervision under both modalities:
\begin{equation}
\label{eq:cause_worst}
\begin{aligned}
\mathcal L_F^i
={}&(1-2\rho)\tau\log\left(
\frac{
e^{\mathcal L_i^{\mathrm{mm}}/\tau}
+
e^{\mathcal L_i^{\mathrm{txt}}/\tau}
}{2}
\right)\\
&+\rho\big(\mathcal L_i^{\mathrm{mm}}
+\mathcal L_i^{\mathrm{txt}}\big).
\end{aligned}
\end{equation}
Here $\tau>0$ controls the sensitivity to modality-specific loss differences, while $0<\rho<1/2$ ensures supervision under both modalities. The resulting gradient weights are $w_i^r=\rho+(1-2\rho)\operatorname{softmax}_r(\mathcal L_i^r/\tau)$, satisfying $\sum_r w_i^r=1$ and $\rho\leq w_i^r\leq1-\rho$, thereby enabling fact-wise coordination of cross-modal suppression.

\subsection{Retain-Constrained Optimization}
\label{sec:cause_retain}

To limit unintended changes to retained knowledge, CAUSE preserves the original predictive distributions and discourages unnecessary uncertainty responses. For a retain input-answer pair $(x,y)$, let $p_0^{x,j}$ and $p_\phi^{x,j}$ denote the original and edited next-token distributions conditioned on $x$ and $y_{<j}$. The distribution-matching objective is
\begin{equation}
\label{eq:cause_retain_kl}
\mathcal K(x,y)
=
\frac{1}{|y|}
\sum_{j=1}^{|y|}
D_{\mathrm{KL}}
\big(p_0^{x,j}\,\|\,p_\phi^{x,j}\big).
\end{equation}
To counteract uncertainty supervision on target facts, CAUSE penalizes excessive predictive support for uncertainty responses on retained examples:
\begin{equation}
\label{eq:cause_retain_response}
\mathcal L_{\mathrm{AR}}(x,y)
=
\left[
\bar s_\phi(x,y_\bot)
-
\operatorname{sg}(\bar s_\phi(x,y))
+
m_R
\right]_+,
\end{equation}
where $m_R>0$ specifies the retention margin. The retention objective integrates these terms across modalities:
\begin{equation}
\label{eq:cause_retain}
\begin{aligned}
\mathcal L_R
=\frac{1}{2}\sum_{r\in\{\mathrm{mm},\mathrm{txt}\}}
\big(
&\lambda_{\mathrm G}\mathcal K_{\mathrm{gold}}^r
+\lambda_{\mathrm S}\mathcal K_{\mathrm{seq}}^r\\
&+\lambda_{\mathrm A}\mathcal K_{\mathrm{aux}}^r
+\lambda_{\mathrm{AR}}\mathcal L_{\mathrm{AR}}^r
\big),
\end{aligned}
\end{equation}
where $\mathcal K_{\mathrm{gold}}^r$, $\mathcal K_{\mathrm{seq}}^r$, and $\mathcal K_{\mathrm{aux}}^r$ match predictive distributions using reference answers, sequences generated by the frozen original model, and auxiliary queries (\eg classification and cloze completion), respectively. The $\lambda$ coefficients balance the corresponding retention terms.

The joint training objective combines factual suppression, retention, and regularization:
\begin{equation}
\label{eq:cause_total}
\min_\phi\quad
\mathbb E_{z_i\sim\mathcal D_F}[\mathcal L_F^i]
+
\mathbb E_{\mathcal D_R}[\mathcal L_R]
+
\lambda_{\mathrm{reg}}\Omega(\phi),
\end{equation}
where $\Omega(\phi)=\sum_{\ell\in\mathcal S}(\|A_\ell\|_F^2+\|B_\ell\|_F^2)/(dk)$ regularizes the shared editors. At each optimization step, let $g_F=\nabla_\phi\mathcal L_F^i$ and $g_R^r=\frac{1}{2}\nabla_\phi\mathcal L_R^r$ denote the forgetting gradient and modality-specific retention gradients, respectively, where $\mathcal L_R^r$ is the corresponding modality-specific loss in Eq.~\eqref{eq:cause_retain}. To limit interference with retained knowledge, CAUSE projects the forgetting gradient as
\begin{equation}
\label{eq:cause_projection}
\begin{aligned}
\tilde g_F &= \arg\min_g \frac{1}{2}\|g-g_F\|_2^2,\\
\text{s.t.}\quad &g^\top g_R^r\geq 0,\quad
r\in\{\mathrm{mm},\mathrm{txt}\}.
\end{aligned}
\end{equation}
This projection mitigates first-order interference between forgetting and retention under both modalities. The resulting gradient $\tilde g_F+\sum_r g_R^r+\lambda_{\mathrm{reg}}\nabla_\phi\Omega(\phi)$ is used to update the shared editors $\phi$, while $\theta_0$ remains frozen.
\section{Experimental Results}
\subsection{Experimental Setup and Implementation Details}

\begin{table*}[t]
\centering
\caption{
Unlearning results on UMU-Bench with LLaVA-1.5-7B
under three forget ratios.
MM and TXT denote multimodal and text-only evaluation,
respectively.
Cls. and Cloze report accuracy (\%), while Gen. reports ROUGE-L.
Mix denotes unlearning with joint multimodal and text-only
supervision.
Gray rows indicate single-modality controls.
The \textbf{best} and \underline{second-best} results
are highlighted, excluding Vanilla and single-modality controls.
}
\label{tab:umu_main_results}

\fontsize{7}{8}\selectfont
\setlength{\tabcolsep}{1.8pt}
\renewcommand{\arraystretch}{1.12}

\begin{adjustbox}{max width=\textwidth}
\begin{tabular*}{\textwidth}{
@{\extracolsep{\fill}}l*{18}{c}@{}
}
\toprule

\textbf{Method}
& \multicolumn{6}{c}{\textbf{Forget-5}}
& \multicolumn{6}{c}{\textbf{Forget-10}}
& \multicolumn{6}{c}{\textbf{Forget-15}}
\\

\cmidrule(lr){2-7}
\cmidrule(lr){8-13}
\cmidrule(lr){14-19}

& \multicolumn{2}{c}{Cls.}
& \multicolumn{2}{c}{Cloze}
& \multicolumn{2}{c}{Gen.}
& \multicolumn{2}{c}{Cls.}
& \multicolumn{2}{c}{Cloze}
& \multicolumn{2}{c}{Gen.}
& \multicolumn{2}{c}{Cls.}
& \multicolumn{2}{c}{Cloze}
& \multicolumn{2}{c}{Gen.}
\\

\cmidrule(lr){2-3}
\cmidrule(lr){4-5}
\cmidrule(lr){6-7}
\cmidrule(lr){8-9}
\cmidrule(lr){10-11}
\cmidrule(lr){12-13}
\cmidrule(lr){14-15}
\cmidrule(lr){16-17}
\cmidrule(lr){18-19}

& MM & TXT & MM & TXT & MM & TXT
& MM & TXT & MM & TXT & MM & TXT
& MM & TXT & MM & TXT & MM & TXT
\\

\midrule


\rowcolor{gray!20}
\multicolumn{19}{l}{\textbf{Forget Set} $\downarrow$}
\\

Vanilla
& 68.00 & 82.00 & 88.00 & 74.00 & .835 & .762
& 71.00 & 72.00 & 92.00 & 69.00 & .779 & .675
& 68.67 & 63.33 & 89.33 & 70.67 & .750 & .712 \\

\addlinespace[1pt]

\rowcolor{gray!9}
Grad. Diff \cite{liu2022continual}
& 56.00 & 76.00 & 36.00 & 70.00 & .337 & .596
& 51.00 & 67.00 & 43.00 & 58.00 & .351 & .648
& 48.00 & 60.00 & 40.00 & 56.67 & .317 & .691 \\

\rowcolor{gray!9}
KL\_Min \cite{nguyen2020variational}
& 58.00 & 78.00 & 38.00 & 70.00 & .362 & .614
& 54.00 & 69.00 & 46.00 & 60.00 & .319 & .623
& 50.00 & 58.00 & 42.67 & 58.67 & .311 & .676 \\

\rowcolor{gray!9}
NPO \cite{zhang2024negative}
& 54.00 & 76.00 & 44.00 & 68.00 & .491 & .713
& 50.00 & 66.00 & 47.00 & 61.00 & .507 & .659
& 47.33 & 56.67 & 49.33 & 61.33 & .493 & .701 \\

\addlinespace[1.5pt]

GD-Mix \cite{liu2022continual,wang2025umu}
& 46.00 & 60.00 & \underline{32.00} & \underline{28.00}
& .335 & .309
& 46.00 & 51.00 & 37.00 & \underline{26.00}
& .294 & \underline{.207}
& 43.33 & 47.33 & 30.00 & 24.67
& .273 & \underline{.228} \\

KL-Mix \cite{nguyen2020variational,wang2025umu}
& 48.00 & 54.00 & 34.00 & 30.00 & .349 & .317
& 47.00 & 52.00 & 39.00 & 28.00 & .304 & .280
& 45.33 & 50.00 & 35.33 & 27.33 & .295 & .311 \\

NPO-Mix \cite{zhang2024negative,wang2025umu}
& 46.00 & 52.00 & 40.00 & 32.00 & .401 & .422
& 43.00 & 40.00 & 41.00 & 31.00 & .336 & .358
& 46.00 & 41.33 & 38.00 & 30.00 & .313 & .334 \\

RMU-Mix \cite{wmdp,wang2025umu}
& 38.00 & \underline{42.00}
& 34.00 & 30.00 & .326 & .298
& 39.00 & 41.00 & 35.00 & 28.00
& \underline{.274} & .266
& 36.67 & 32.00 & 31.33 & 26.67
& \underline{.261} & .248 \\

\addlinespace[1.5pt]

MANU \cite{liu2025modality}
& 60.00 & 72.00 & 70.00 & 66.00 & .682 & .613
& 60.00 & 59.00 & 69.00 & 61.00 & .637 & .602
& 57.33 & 54.67 & 64.00 & 61.33 & .619 & .624 \\

M.E. \cite{li2026crossmodal}
& \underline{32.00} & 44.00
& 40.00 & 40.00 & \underline{.310} & .299
& \underline{32.00} & \textbf{31.00}
& 42.00 & 38.00 & .306 & .293
& \underline{30.67} & \textbf{26.67}
& 38.67 & 35.33 & .295 & .273 \\

CMJU \cite{yang2026cmju}
& 40.00 & 44.00
& 38.00 & 36.00 & .329 & \underline{.283}
& 36.00 & \underline{32.00}
& \underline{33.00} & 29.00 & .311 & .215
& 31.33 & 29.33
& \underline{26.67} & \underline{21.33}
& .307 & .251 \\

\addlinespace[1pt]

\rowcolor{blue!8}
\textbf{CAUSE (Ours)}
& \textbf{28.00} & \textbf{34.00}
& \textbf{14.00} & \textbf{20.00}
& \textbf{.036} & \textbf{.057}
& \textbf{24.00} & \textbf{31.00}
& \textbf{12.00} & \textbf{18.00}
& \textbf{.045} & \textbf{.062}
& \textbf{21.33} & \underline{27.33}
& \textbf{11.33} & \textbf{16.67}
& \textbf{.027} & \textbf{.031} \\

\midrule


\rowcolor{gray!20}
\multicolumn{19}{l}{\textbf{Retain Set} $\uparrow$}
\\

Vanilla
& 69.47 & 66.95 & 89.58 & 70.84 & .773 & .717
& 69.22 & 67.22 & 89.22 & 71.22 & .776 & .724
& 69.53 & 68.47 & 89.53 & 71.06 & .781 & .721 \\

\addlinespace[1pt]

\rowcolor{gray!9}
Grad. Diff \cite{liu2022continual}
& 59.26 & 64.84 & 46.32 & 67.26 & .429 & .685
& 54.33 & 62.78 & 43.56 & 65.89 & .403 & .652
& 51.76 & 64.35 & 44.94 & 61.41 & .345 & .637 \\

\rowcolor{gray!9}
KL\_Min \cite{nguyen2020variational}
& 60.00 & 64.95 & 48.42 & 68.00 & .445 & .690
& 57.11 & 64.78 & 47.22 & 64.78 & .419 & .662
& 54.94 & 65.18 & 46.82 & 64.00 & .409 & .658 \\

\rowcolor{gray!9}
NPO \cite{zhang2024negative}
& 59.89 & 62.84 & 54.00 & 70.00 & .571 & .712
& 56.44 & 63.44 & 54.78 & 66.56 & .557 & .690
& 53.41 & 62.94 & 51.06 & 64.71 & .533 & .694 \\

\addlinespace[1.5pt]

GD-Mix \cite{liu2022continual,wang2025umu}
& 46.53 & 52.11 & 40.11 & 32.95 & .337 & .294
& 43.78 & 50.33 & 39.78 & 33.44 & .311 & .275
& 44.24 & 46.24 & 37.76 & 29.76 & .299 & .262 \\

KL-Mix \cite{nguyen2020variational,wang2025umu}
& 47.89 & 50.42 & 42.95 & 39.05 & .394 & .342
& 45.89 & 48.78 & 41.00 & 34.33 & .353 & .326
& 46.94 & 42.35 & 39.41 & 32.00 & .318 & .287 \\

NPO-Mix \cite{zhang2024negative,wang2025umu}
& 51.16 & 53.26 & 45.05 & 40.11 & .458 & .473
& 48.33 & 46.11 & 44.00 & 39.89 & .389 & .366
& 49.18 & 44.47 & 41.53 & 36.82 & .372 & .309 \\

RMU-Mix \cite{wmdp,wang2025umu}
& 45.68 & 31.47 & 41.47 & 33.58 & .362 & .317
& 41.78 & 39.22 & 39.67 & 34.22 & .320 & .275
& 38.59 & 36.71 & 40.12 & 32.47 & .308 & .268 \\

\addlinespace[1.5pt]

MANU \cite{liu2025modality}
& \underline{63.37} & \underline{59.16}
& \underline{70.42} & \textbf{65.26}
& \textbf{.662} & \textbf{.597}
& \underline{58.33} & \underline{57.33}
& \underline{64.33} & \textbf{63.11}
& \textbf{.645} & \textbf{.614}
& \underline{58.47} & \underline{56.35}
& \underline{65.06} & \underline{62.35}
& \textbf{.628} & \textbf{.637} \\

M.E. \cite{li2026crossmodal}
& 33.79 & 31.37 & 42.42 & 36.21 & .324 & .318
& 30.89 & 29.78 & 41.89 & 38.22 & .313 & .302
& 31.76 & 30.24 & 40.00 & 36.00 & .340 & .315 \\

CMJU \cite{yang2026cmju}
& 50.53 & 47.79 & 60.84 & 52.74 & .469 & .447
& 48.33 & 46.22 & 58.33 & 51.00 & .446 & .419
& 47.53 & 45.29 & 56.12 & 48.59 & .452 & .405 \\

\addlinespace[1pt]

\rowcolor{blue!8}
\textbf{CAUSE (Ours)}
& \textbf{65.58} & \textbf{62.63}
& \textbf{82.32} & \underline{63.47}
& \underline{.588} & \underline{.561}
& \textbf{63.33} & \textbf{60.11}
& \textbf{80.89} & \underline{61.11}
& \underline{.592} & \underline{.543}
& \textbf{61.76} & \textbf{62.35}
& \textbf{77.88} & \textbf{64.24}
& \underline{.579} & \underline{.553} \\

\bottomrule
\end{tabular*}
\end{adjustbox}
\end{table*}

To comprehensively assess the effectiveness, selectivity, and cross-modal consistency of factual unlearning, we conduct experiments on two dedicated multimodal unlearning benchmarks: UMU-Bench \cite{wang2025umu} and MLLMU-Bench \cite{liu2025protecting}. MLLMU-Bench enables us to evaluate forgetting effectiveness and utility preservation across diverse multimodal scenarios, while UMU-Bench explicitly examines whether targeted knowledge remains recoverable through alternative modality access routes. Together, these benchmarks allow us to assess not only whether targeted knowledge is suppressed, but also whether such suppression is consistent across modalities and selective with respect to non-target knowledge. We evaluate LLaVA-1.5-7B \cite{liu2024improved} and Idefics2-8B \cite{laurenccon2024matters} on MLLMU-Bench, and LLaVA-1.5-7B on UMU-Bench.

\subsection{Baseline Methods}
We compare CAUSE with seven unlearning methods: Gradient Difference (Grad. Diff) \cite{liu2022continual}, KL Minimization (KL\_Min) \cite{nguyen2020variational}, Negative Preference Optimization (NPO) \cite{zhang2024negative}, Representation Misdirection for Unlearning (RMU) \cite{wmdp}, MANU \cite{liu2025modality}, MIP-Editor (M.E.) \cite{li2026crossmodal}, and Cross-Modal Joint Unlearning (CMJU) \cite{yang2026cmju}. The former four are representative LLM unlearning methods, while the latter three are specifically designed for MLLM unlearning. Following the joint multimodal and text-only supervision setting in UMU-Bench \cite{wang2025umu}, we evaluate the corresponding Mix variants, namely GD-Mix, KL-Mix, NPO-Mix, and RMU-Mix. The single-modality variants of Grad. Diff, KL\_Min, and NPO serve as additional controls to examine the effect of expanding unlearning supervision across input modalities. 

\subsection{Evaluation Results on UMU-Bench}

Table~\ref{tab:umu_main_results} presents the forgetting and retention performance of different unlearning methods on UMU-Bench across three forget ratios. Conventional LLM unlearning methods, including Grad.~Diff., KL\_Min, and NPO, exhibit substantial cross-modal disparities, suppressing factual predictions under multimodal inputs while leaving considerable target knowledge accessible through text-only queries. These results demonstrate the limited transferability of forgetting across modalities.

Joint multimodal and text-only supervision improves forgetting and reduces cross-modal disparities, as demonstrated by GD-Mix, KL-Mix, and NPO-Mix~\cite{wang2025umu}, but incurs considerable retention degradation. Neuron-based methods exhibit contrasting behaviors: MANU achieves relatively strong retention but limited forgetting, whereas M.E. achieves stronger suppression at the expense of model utility, suggesting that identifying knowledge-related neurons alone is insufficient to determine an appropriate intervention strength. CMJU improves this trade-off through modality-aware gradient coordination. CAUSE achieves either the best or second-best result on each evaluated metric across all three forget ratios, excluding the single-modality controls. RMU-Mix provides a particularly relevant baseline, as it also performs unlearning at the representation level. Compared with RMU-Mix, CAUSE achieves stronger factual suppression and substantially better retention, with particularly pronounced improvements in generation across both modalities. These results suggest the effectiveness of context-anchored state editing with calibrated suppression and retention constraints. Increasing the forget ratio reveals different scaling behaviors across methods. Conventional LLM unlearning methods generally achieve stronger multimodal forgetting, whereas text-only forgetting improves inconsistently, particularly in generation. The Mix variants and CMJU exhibit stronger factual suppression across modalities but declining retention. In contrast, CAUSE achieves stronger overall factual suppression while maintaining relatively stable retention as the forget ratio increases from 5\% to 15\%. With a fixed 1:1 forget-to-retain sampling ratio for methods using retain supervision, the amount of retain supervision scales proportionally with forget supervision. These results highlight the ability of CAUSE to maintain a favorable forgetting-retention trade-off across different unlearning scales. 

\subsection{Fact-Level Cross-Modal Consistency}

\begin{figure}
        \includegraphics[width=0.45\textwidth,height=0.28\textwidth]{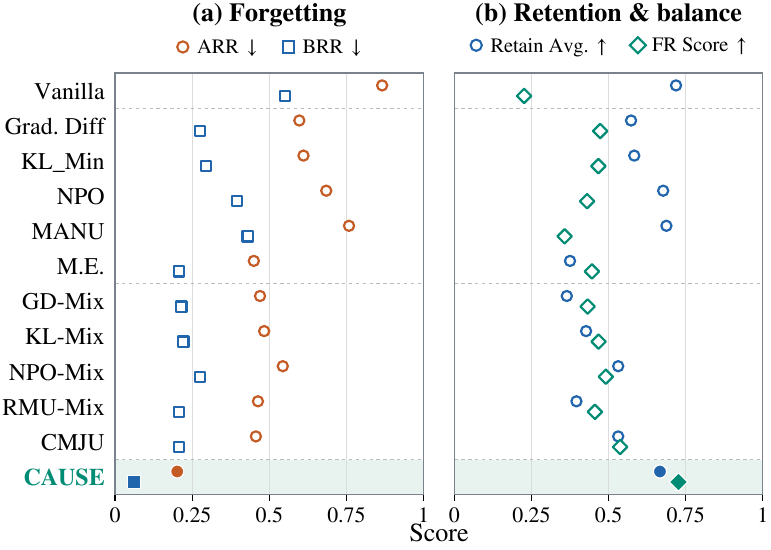}
        \caption{\label{fig:fc}\small Fact-level cross-modal forgetting and retention on UMU-Bench (Forget-5), evaluated via generation-based lexical recovery on 149 forget facts and 48 retain facts. (a) ARR and BRR measure factual recovery through either or both modalities, respectively. (b) Retain Avg. measures mean recovery across modalities, and FR Score is the harmonic mean of $1-\mathrm{ARR}$ and Retain Avg.}
\end{figure}

To further assess cross-modal forgetting consistency, we conduct a fact-level analysis on UMU-Bench (Forget-5), examining whether individual target facts can be recovered through either or both input modalities, as illustrated in Fig.~\ref{fig:fc}. Conventional unlearning methods exhibit substantial residual factual recovery, while the Mix variants and CMJU improve cross-modal suppression but often compromise retention. CAUSE achieves the lowest ARR and BRR while maintaining strong retention, resulting in the highest FR Score. Although NPO and MANU exhibit slightly higher retention rates, their substantially higher factual recovery rates indicate limited forgetting effectiveness. Compared with CMJU, CAUSE achieves stronger cross-modal factual suppression and better non-target knowledge preservation. These results demonstrate the effectiveness of CAUSE in promoting consistent factual unlearning across modalities while preserving retained knowledge.

\begin{figure}
        \includegraphics[width=0.45\textwidth]{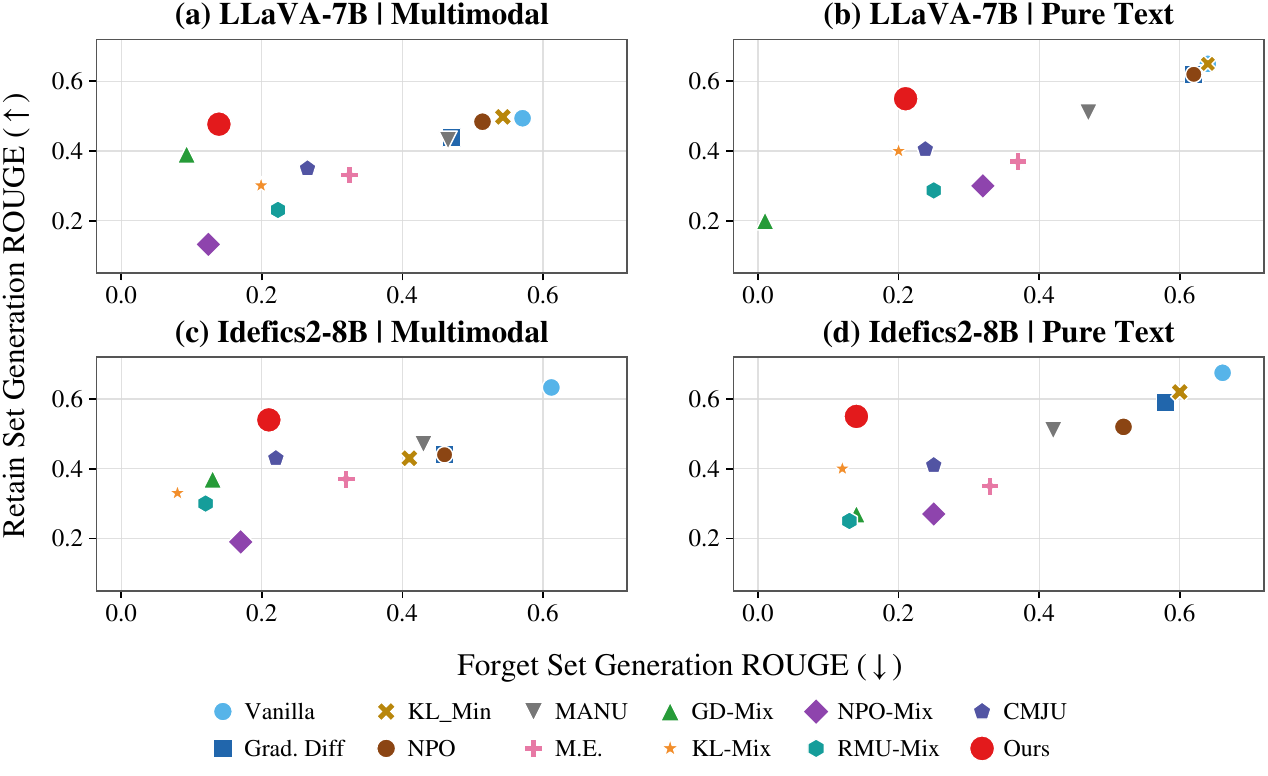}
        \caption{\label{fig:mllmu}\small \small Forget-retain trade-off in generation performance on MLLMU-Bench at a 5\% forgetting ratio for LLaVA-7B and Idefics2-8B across both query modalities. The upper-left region represents a more favorable unlearning-utility trade-off.}
\end{figure}

\begin{table}[b]
\centering
\caption{\small Component-wise ablation of CAUSE on a fact-level diagnostic set from UMU-Bench (Forget-5), comprising 149 forget facts and 48 retain facts. Evaluation is based on generation-based lexical recovery. ARR, BRR, and RD denote any-route recovery rate, both-route recovery rate, and route discordance, respectively. Avg. denotes the mean recovery rate across modalities on the retain set.}
\label{tab:ablation}

\scriptsize
\setlength{\tabcolsep}{3.5pt}
\renewcommand{\arraystretch}{1.12}

\begin{tabular*}{\columnwidth}
{@{\extracolsep{\fill}}lcccc@{}}
\toprule
\multirow{2}{*}{Method}
& \multicolumn{3}{c}{Forget Set}
& Retain Set \\
\cmidrule(lr){2-4}
\cmidrule(lr){5-5}
& ARR $\downarrow$
& BRR $\downarrow$
& RD $\downarrow$
& Avg. $\uparrow$ \\
\midrule
w/o Context Anchoring
    & 0.215 & 0.067 & 0.148 & 0.583 \\
w/o Shared Editing
    & 0.228 & 0.074 & 0.154 & 0.646 \\
\midrule
w/o Identity Calibration
    & 0.262 & 0.087 & 0.174 & 0.635 \\
w/o Bounded Suppression
    & 0.174 & 0.034 & 0.141 & 0.625 \\
w/o Confidence Suppression
    & 0.221 & 0.067 & 0.154 & 0.677 \\
\makecell[l]{w/o Adaptive Cross-Modal\\Suppression}
    & 0.349 & 0.134 & 0.215 & 0.677 \\
\midrule
w/o Retain Constraints
    & 0.114 & 0.027 & 0.087 & 0.479 \\
\midrule
\rowcolor{blue!8}
\textbf{CAUSE (Full)}
    & 0.201 & 0.060 & 0.141 & 0.667 \\
\bottomrule
\end{tabular*}
\end{table}

\subsection{Generalization across Models and Benchmarks}

Fig.~\ref{fig:mllmu} presents the forgetting-retention trade-off in generation performance on MLLMU-Bench for LLaVA-7B and Idefics2-8B at a forget ratio of 5\%. Across both models, conventional LLM unlearning methods exhibit limited forgetting, particularly under pure-text queries. The Mix variants achieve stronger factual suppression but impose substantial retention losses. CAUSE achieves a favorable trade-off across both model architectures and input modalities, maintaining high retention performance while substantially reducing forget-set ROUGE-L. The relative forgetting performance across input modalities varies between the two model architectures. CAUSE achieves lower forget-set ROUGE-L under multimodal queries on LLaVA-7B, whereas the opposite pattern is observed on Idefics2-8B. This variation indicates that the relative suppression requirements of different input modalities may depend on the underlying model. Nevertheless, CAUSE maintains a substantial separation between forget and retain performance across all four settings. These results demonstrate the applicability of CAUSE beyond a specific MLLM architecture. 
\begin{figure}
        \includegraphics[width=0.48\textwidth]{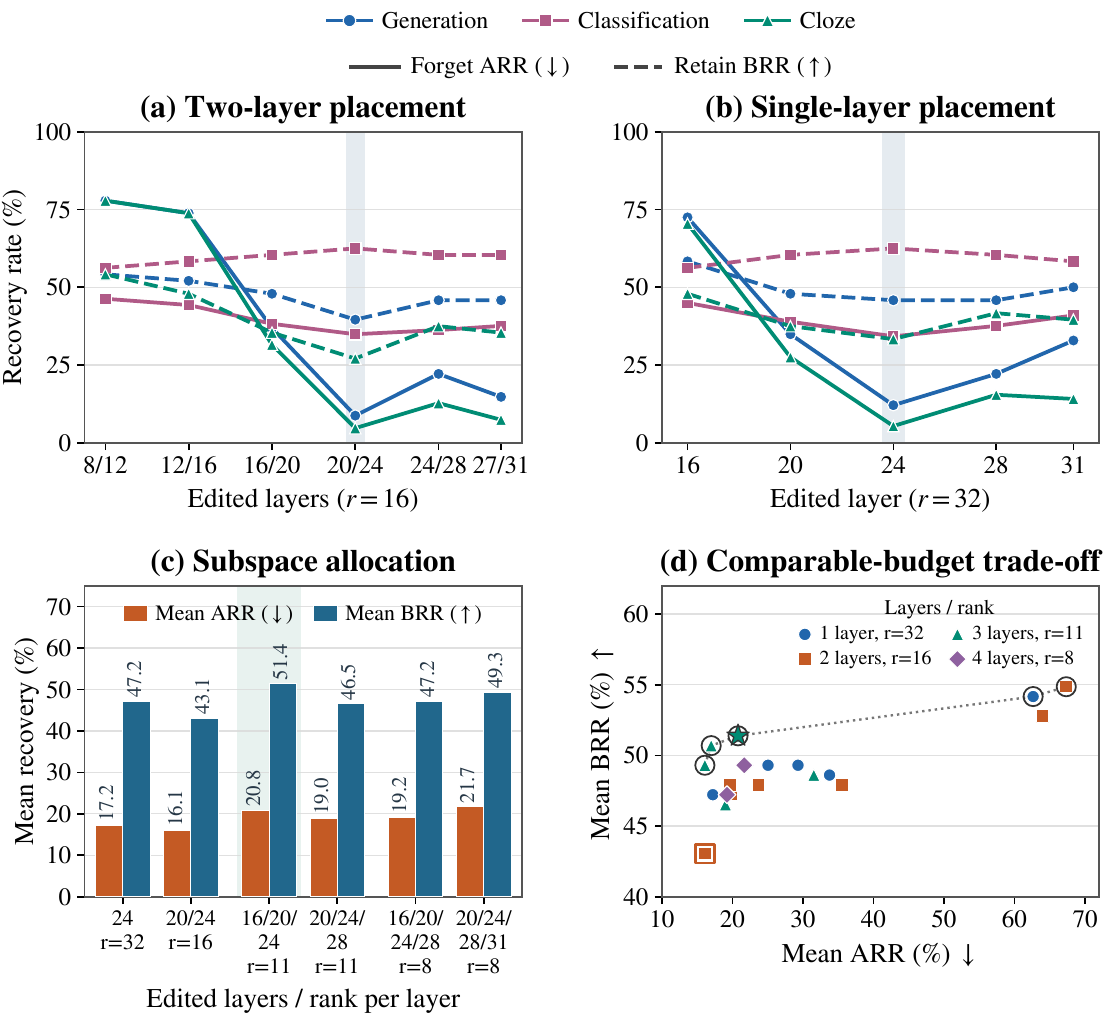}
        \caption{\small Effects of intervention placement and editing capacity on the forgetting-retention trade-off on UMU-Bench (Forget-5), evaluated on 149 forget facts and 48 retain facts. Mean ARR and Mean BRR denote the forget-set ARR and retain-set BRR averaged across generation, classification, and cloze completion, respectively. 
\label{fig:layer_analysis}}
\vspace{-3mm}
\end{figure}
\subsection{Ablation Studies}

Table~\ref{tab:ablation} and Fig.~\ref{fig:layer_analysis} examine the contributions of individual components and the effects of intervention placement in CAUSE. Context anchoring and shared editing improve factual suppression and reduce cross-modal discordance, with context anchoring offering substantial retention benefits. Identity calibration improves both forgetting and retention, while confidence suppression reduces residual factual recovery. Bounded suppression and retain constraints improve knowledge retention by limiting excessive suppression and interference with non-target predictions, respectively. Removing adaptive cross-modal suppression increases ARR and RD, highlighting the importance of fact-wise coordination across modalities. The layer-wise analysis reveals that forgetting effectiveness varies with intervention depth, particularly for generation and cloze completion. Fig.~\ref{fig:layer_analysis}(c) further examines the allocation of editing capacity across layers and ranks, while Fig.~\ref{fig:layer_analysis}(d) summarizes the forgetting-retention trade-off across tasks. Under comparable parameter budgets, distributing editing capacity across selected middle-to-late decoder layers provides a favorable balance between factual suppression and knowledge retention.

\begin{figure}
\centering
        \includegraphics[width=0.35\textwidth, height=0.2\textwidth]{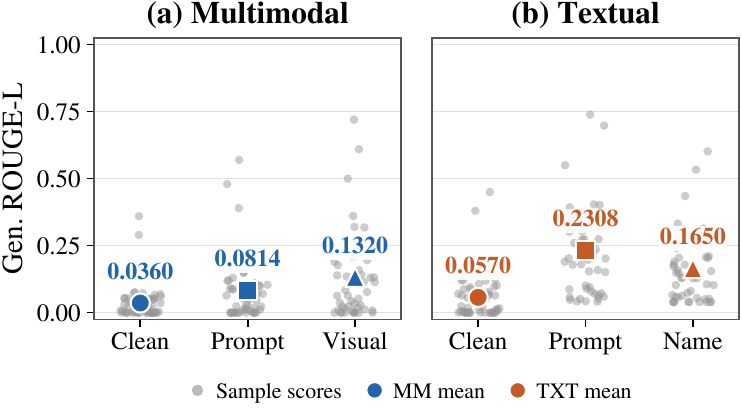}
        \caption{\small Robustness of CAUSE against adversarial knowledge recovery. Both routes are evaluated using clean queries and adversary prompt attacks, with additional visual perturbations for MM and name variations for TXT. Gray dots denote sample scores, and colored markers indicate means. Lower ROUGE-L indicates stronger factual suppression.
\label{fig:robut}}
\end{figure}
\subsection{Robustness against Adversarial Recovery}

We evaluate the robustness of CAUSE on 50 paired forget samples under multimodal and textual input variations. As illustrated in Fig.~\ref{fig:robut}, CAUSE achieves mean ROUGE-L scores of 0.081 and 0.231 under multimodal and textual prompt attacks, respectively, while visual perturbations and name variations yield scores of 0.132 and 0.165. Although textual prompt attacks induce the largest increase in ROUGE-L, input variations may elicit alternative factual responses rather than uncertainty expressions, increasing lexical overlap with reference answers without necessarily recovering the original target facts. These results suggest that CAUSE exhibits robustness to diverse input variations beyond the original query formulations. 

\section{Conclusion}
\label{sec:conclusion}

In this work, we investigate modality-inconsistent forgetting in MLLMs and reveal that multimodal and text-only queries referring to the same factual knowledge induce distinct contextual representations and undergo asymmetric changes during multimodal-only unlearning. Motivated by these observations, we propose CAUSE, a multimodal unlearning framework that establishes the context-to-answer transition as a shared intervention interface across modalities. CAUSE calibrates factual suppression using identity-related differences in predictive support, adaptively coordinates modality-specific forgetting objectives, and incorporates retention constraints to preserve non-target knowledge. Extensive experiments on MLLMU-Bench and UMU-Bench demonstrate that CAUSE achieves more consistent unlearning across modalities while maintaining a favorable forgetting-retention trade-off. Our findings underscore the importance of jointly considering intervention placement, modality-specific suppression requirements, and knowledge preservation in selective multimodal unlearning.

{
    \small
    \bibliographystyle{ieeenat_fullname}
    \bibliography{main}

\begin{thebibliography}{33}
\providecommand{\natexlab}[1]{#1}
\providecommand{\url}[1]{\texttt{#1}}
\expandafter\ifx\csname urlstyle\endcsname\relax
  \providecommand{\doi}[1]{doi: #1}\else
  \providecommand{\doi}{doi: \begingroup \urlstyle{rm}\Url}\fi

\bibitem[Brown et~al.(2020)Brown, Mann, Ryder, Subbiah, Kaplan, Dhariwal, Neelakantan, Shyam, Sastry, Askell, et~al.]{brown2020language}
Tom Brown, Benjamin Mann, Nick Ryder, Melanie Subbiah, Jared~D Kaplan, Prafulla Dhariwal, Arvind Neelakantan, Pranav Shyam, Girish Sastry, Amanda Askell, et~al.
\newblock Language models are few-shot learners.
\newblock \emph{Advances in neural information processing systems}, 33:\penalty0 1877--1901, 2020.

\bibitem[Chen et~al.(2025)Chen, Li, Zhou, Chen, and Wei]{chen2025unveiling}
Tiejin Chen, Pingzhi Li, Kaixiong Zhou, Tianlong Chen, and Hua Wei.
\newblock Unveiling privacy risks in multi-modal large language models: Task-specific vulnerabilities and mitigation challenges.
\newblock In \emph{Findings of the Association for Computational Linguistics: ACL 2025}, pages 4573--4586, 2025.

\bibitem[Cheng and Amiri(2024)]{cheng2024multidelete}
Jiali Cheng and Hadi Amiri.
\newblock Multidelete for multimodal machine unlearning.
\newblock In \emph{European Conference on Computer Vision}, pages 165--184. Springer, 2024.

\bibitem[Chowdhery et~al.(2023)Chowdhery, Narang, Devlin, Bosma, Mishra, Roberts, Barham, Chung, Sutton, Gehrmann, et~al.]{chowdhery2023palm}
Aakanksha Chowdhery, Sharan Narang, Jacob Devlin, Maarten Bosma, Gaurav Mishra, Adam Roberts, Paul Barham, Hyung~Won Chung, Charles Sutton, Sebastian Gehrmann, et~al.
\newblock Palm: Scaling language modeling with pathways.
\newblock \emph{Journal of Machine Learning Research}, 24\penalty0 (240):\penalty0 1--113, 2023.

\bibitem[Dong et~al.(2025)Dong, Zhou, Liu, Yang, and Lu]{dong2025emergent}
Zhichen Dong, Zhanhui Zhou, Zhixuan Liu, Chao Yang, and Chaochao Lu.
\newblock Emergent response planning in {LLM}s.
\newblock In \emph{Proceedings of the 42nd International Conference on Machine Learning}, pages 14301--14320. PMLR, 2025.

\bibitem[Dontsov et~al.(2025)Dontsov, Korzh, Zhavoronkin, Mikheev, Bobkov, Alanov, Rogov, Oseledets, and Tutubalina]{dontsov2025clear}
Alexey Dontsov, Dmitrii Korzh, Alexey Zhavoronkin, Boris Mikheev, Denis Bobkov, Aibek Alanov, Oleg Rogov, Ivan Oseledets, and Elena Tutubalina.
\newblock Clear: Character unlearning in textual and visual modalities.
\newblock In \emph{Findings of the Association for Computational Linguistics: ACL 2025}, pages 20582--20603, 2025.

\bibitem[Fan et~al.(2024)Fan, Liu, Zhang, Wong, Wei, and Liu]{fan2024salun}
Chongyu Fan, Jiancheng Liu, Yihua Zhang, Eric Wong, Dennis Wei, and Sijia Liu.
\newblock Salun: Empowering machine unlearning via gradient-based weight saliency in both image classification and generation.
\newblock In \emph{International Conference on Learning Representations}, pages 53643--53673, 2024.

\bibitem[Fan et~al.(2025)Fan, Liu, Lin, Jia, Zhang, Mei, and Liu]{NEURIPS2025_02443e4e}
Chongyu Fan, Jiancheng Liu, Licong Lin, Jinghan Jia, Ruiqi Zhang, Song Mei, and Sijia Liu.
\newblock Simplicity prevails: Rethinking negative preference optimization for llm unlearning.
\newblock In \emph{Advances in Neural Information Processing Systems}, pages 1540--1567, 2025.

\bibitem[Hong et~al.(2025)Hong, Zhao, Tang, Deng, Rong, and Zhang]{NEURIPS2025_8e8388ec}
Yihuai Hong, Yiran Zhao, Wei Tang, Yang Deng, Yu Rong, and Wenxuan Zhang.
\newblock The rise of parameter specialization for knowledge storage in large language models.
\newblock In \emph{Advances in Neural Information Processing Systems}, pages 98940--98965, 2025.

\bibitem[Huo et~al.(2025)Huo, Yan, Zheng, Lyu, Zou, Wei, and Hu]{huo2025mmunlearner}
Jiahao Huo, Yibo Yan, Xu Zheng, Yuanhuiyi Lyu, Xin Zou, Zhihua Wei, and Xuming Hu.
\newblock Mmunlearner: Reformulating multimodal machine unlearning in the era of multimodal large language models.
\newblock In \emph{Findings of the Association for Computational Linguistics: ACL 2025}, pages 7190--7206, 2025.

\bibitem[Jang et~al.(2023)Jang, Yoon, Yang, Cha, Lee, Logeswaran, and Seo]{jang-etal-2023-knowledge}
Joel Jang, Dongkeun Yoon, Sohee Yang, Sungmin Cha, Moontae Lee, Lajanugen Logeswaran, and Minjoon Seo.
\newblock Knowledge unlearning for mitigating privacy risks in language models.
\newblock In \emph{Proceedings of the 61st Annual Meeting of the Association for Computational Linguistics (Volume 1: Long Papers)}. Association for Computational Linguistics, 2023.

\bibitem[Kandpal et~al.(2022)Kandpal, Wallace, and Raffel]{kandpal2022deduplicating}
Nikhil Kandpal, Eric Wallace, and Colin Raffel.
\newblock Deduplicating training data mitigates privacy risks in language models.
\newblock In \emph{International Conference on Machine Learning}, pages 10697--10707. PMLR, 2022.

\bibitem[Lauren{\c{c}}on et~al.(2024)Lauren{\c{c}}on, Tronchon, Cord, and Sanh]{laurenccon2024matters}
Hugo Lauren{\c{c}}on, L{\'e}o Tronchon, Matthieu Cord, and Victor Sanh.
\newblock What matters when building vision-language models?
\newblock \emph{Advances in Neural Information Processing Systems}, 37:\penalty0 87874--87907, 2024.

\bibitem[Li et~al.(2023)Li, Li, Savarese, and Hoi]{li2023blip}
Junnan Li, Dongxu Li, Silvio Savarese, and Steven Hoi.
\newblock Blip-2: Bootstrapping language-image pre-training with frozen image encoders and large language models.
\newblock In \emph{International conference on machine learning}, pages 19730--19742. PMLR, 2023.

\bibitem[Li et~al.(2024{\natexlab{a}})Li, Wei, Zhang, Qi, Du, Chen, Bi, and Liu]{li2024single}
Jiaqi Li, Qianshan Wei, Chuanyi Zhang, Guilin Qi, Miaozeng Du, Yongrui Chen, Sheng Bi, and Fan Liu.
\newblock Single image unlearning: Efficient machine unlearning in multimodal large language models.
\newblock \emph{Advances in Neural Information Processing Systems}, 37:\penalty0 35414--35453, 2024{\natexlab{a}}.

\bibitem[Li et~al.(2026)Li, Li, Wu, Yang, Bai, Jia, and Xue]{li2026crossmodal}
Kunhao Li, Wenhao Li, Di Wu, Lei Yang, Jun Bai, Ju Jia, and Jason Xue.
\newblock Cross-modal unlearning via influential neuron path editing in multimodal large language models.
\newblock In \emph{Proceedings of the AAAI Conference on Artificial Intelligence (AAAI)}, 2026.

\bibitem[Li et~al.(2024{\natexlab{b}})Li, Pan, Gopal, Yue, Berrios, Gatti, Li, Dombrowski, Goel, Mukobi, Helm-Burger, Lababidi, Justen, Liu, Chen, Barrass, Zhang, Zhu, Tamirisa, Bharathi, Herbert-Voss, Breuer, Zou, Mazeika, Wang, Oswal, Lin, Hunt, Tienken-Harder, Shih, Talley, Guan, Steneker, Campbell, Jokubaitis, Basart, Fitz, Kumaraguru, Karmakar, Tupakula, Varadharajan, Shoshitaishvili, Ba, Esvelt, Wang, and Hendrycks]{wmdp}
Nathaniel Li, Alexander Pan, Anjali Gopal, Summer Yue, Daniel Berrios, Alice Gatti, Justin~D. Li, Ann-Kathrin Dombrowski, Shashwat Goel, Gabriel Mukobi, Nathan Helm-Burger, Rassin Lababidi, Lennart Justen, Andrew~Bo Liu, Michael Chen, Isabelle Barrass, Oliver Zhang, Xiaoyuan Zhu, Rishub Tamirisa, Bhrugu Bharathi, Ariel Herbert-Voss, Cort~B Breuer, Andy Zou, Mantas Mazeika, Zifan Wang, Palash Oswal, Weiran Lin, Adam~Alfred Hunt, Justin Tienken-Harder, Kevin~Y. Shih, Kemper Talley, John Guan, Ian Steneker, David Campbell, Brad Jokubaitis, Steven Basart, Stephen Fitz, Ponnurangam Kumaraguru, Kallol~Krishna Karmakar, Uday Tupakula, Vijay Varadharajan, Yan Shoshitaishvili, Jimmy Ba, Kevin~M. Esvelt, Alexandr Wang, and Dan Hendrycks.
\newblock The {WMDP} benchmark: Measuring and reducing malicious use with unlearning.
\newblock In \emph{Proceedings of the 41st International Conference on Machine Learning}, pages 28525--28550. PMLR, 2024{\natexlab{b}}.

\bibitem[Liu et~al.(2022)Liu, Liu, and Stone]{liu2022continual}
Bo Liu, Qiang Liu, and Peter Stone.
\newblock Continual learning and private unlearning.
\newblock In \emph{Conference on Lifelong Learning Agents}, pages 243--254. PMLR, 2022.

\bibitem[Liu et~al.(2024)Liu, Li, Li, and Lee]{liu2024improved}
Haotian Liu, Chunyuan Li, Yuheng Li, and Yong~Jae Lee.
\newblock Improved baselines with visual instruction tuning.
\newblock In \emph{Proceedings of the IEEE/CVF conference on computer vision and pattern recognition}, pages 26296--26306, 2024.

\bibitem[Liu et~al.(2025{\natexlab{a}})Liu, Dou, Jia, Tan, Zeng, Yuan, and Jiang]{liu2025protecting}
Zheyuan Liu, Guangyao Dou, Mengzhao Jia, Zhaoxuan Tan, Qingkai Zeng, Yongle Yuan, and Meng Jiang.
\newblock Protecting privacy in multimodal large language models with mllmu-bench.
\newblock In \emph{Proceedings of the 2025 Conference of the Nations of the Americas Chapter of the Association for Computational Linguistics: Human Language Technologies (Volume 1: Long Papers)}, pages 4105--4135, 2025{\natexlab{a}}.

\bibitem[Liu et~al.(2025{\natexlab{b}})Liu, Dou, Yuan, Zhang, Tan, and Jiang]{liu2025modality}
Zheyuan Liu, Guangyao Dou, Xiangchi Yuan, Chunhui Zhang, Zhaoxuan Tan, and Meng Jiang.
\newblock Modality-aware neuron pruning for unlearning in multimodal large language models.
\newblock In \emph{Proceedings of the 63rd Annual Meeting of the Association for Computational Linguistics (Volume 1: Long Papers)}, pages 5913--5933, 2025{\natexlab{b}}.

\bibitem[Nasr et~al.(2025)Nasr, Rando, Carlini, Hayase, Jagielski, Cooper, Ippolito, Choquette-Choo, Tramer, and Lee]{ICLR2025_cce0e917}
Milad Nasr, Javier Rando, Nicholas Carlini, Jonathan Hayase, Matthew Jagielski, A.~Feder Cooper, Daphne Ippolito, Christopher Choquette-Choo, Florian Tramer, and Katherine Lee.
\newblock Scalable extraction of training data from aligned, production language models.
\newblock In \emph{International Conference on Learning Representations}, pages 82363--82435, 2025.

\bibitem[Nguyen et~al.(2020)Nguyen, Low, and Jaillet]{nguyen2020variational}
Quoc~Phong Nguyen, Bryan Kian~Hsiang Low, and Patrick Jaillet.
\newblock Variational bayesian unlearning.
\newblock \emph{Advances in Neural Information Processing Systems}, 33:\penalty0 16025--16036, 2020.

\bibitem[Ouyang et~al.(2022)Ouyang, Wu, Jiang, Almeida, Wainwright, Mishkin, Zhang, Agarwal, Slama, Ray, et~al.]{ouyang2022training}
Long Ouyang, Jeffrey Wu, Xu Jiang, Diogo Almeida, Carroll Wainwright, Pamela Mishkin, Chong Zhang, Sandhini Agarwal, Katarina Slama, Alex Ray, et~al.
\newblock Training language models to follow instructions with human feedback.
\newblock \emph{Advances in neural information processing systems}, 35:\penalty0 27730--27744, 2022.

\bibitem[Shi et~al.(2025)Shi, Lee, Huang, Malladi, Zhao, Holtzman, Liu, Zettlemoyer, Smith, and Zhang]{shi2025muse}
Weijia Shi, Jaechan Lee, Yangsibo Huang, Sadhika Malladi, Jieyu Zhao, Ari Holtzman, Daogao Liu, Luke Zettlemoyer, Noah Smith, and Chiyuan Zhang.
\newblock Muse: Machine unlearning six-way evaluation for language models.
\newblock In \emph{International Conference on Learning Representations}, pages 27797--27818, 2025.

\bibitem[Tan et~al.(2026)Tan, Li, Cui, Qu, Chen, and Gao]{tan2026geometric}
Chenchen Tan, Xinghao Li, Shujie Cui, Youyang Qu, Cunjian Chen, and Longxiang Gao.
\newblock Less is more: Geometric unlearning for {LLM}s with minimal data disclosure.
\newblock In \emph{Proceedings of the 43rd International Conference on Machine Learning}, pages 118567--118587. PMLR, 2026.

\bibitem[T{\"o}mek{\c{c}}e et~al.(2024)T{\"o}mek{\c{c}}e, Vero, Staab, and Vechev]{tomekcce2024private}
Batuhan T{\"o}mek{\c{c}}e, Mark Vero, Robin Staab, and Martin Vechev.
\newblock Private attribute inference from images with vision-language models.
\newblock \emph{Advances in Neural Information Processing Systems}, 37:\penalty0 103619--103651, 2024.

\bibitem[Wang et~al.(2025)Wang, Li, Feng, Chen, Zheng, and Yin]{wang2025umu}
Chengye Wang, Yuyuan Li, XiaoHua Feng, Chaochao Chen, Xiaolin Zheng, and Jianwei Yin.
\newblock Umu-bench: Closing the modality gap in multimodal unlearning evaluation.
\newblock In \emph{The Thirty-ninth Annual Conference on Neural Information Processing Systems Datasets and Benchmarks Track}, 2025.

\bibitem[Yang et~al.(2026)Yang, Xiong, and Yang]{yang2026cmju}
Di Yang, Baochen Xiong, and Xiaoshan Yang.
\newblock Cmju: cross-modal joint unlearning for balanced forgetting in multimodal large language models.
\newblock \emph{Visual Intelligence}, 4:\penalty0 26, 2026.

\bibitem[Yin et~al.(2025)Yin, Ren, Yan, Ding, and Hao]{yin2025rod}
Heng Yin, Yuqiang Ren, Ke Yan, Shouhong Ding, and Yongtao Hao.
\newblock Rod-mllm: Towards more reliable object detection in multimodal large language models.
\newblock In \emph{Proceedings of the Computer Vision and Pattern Recognition Conference}, pages 14358--14368, 2025.

\bibitem[Yuan et~al.(2025)Yuan, Pang, Du, Chen, Zhang, and Lin]{yuancloser}
Xiaojian Yuan, Tianyu Pang, Chao Du, Kejiang Chen, Weiming Zhang, and Min Lin.
\newblock A closer look at machine unlearning for large language models.
\newblock In \emph{The Thirteenth International Conference on Learning Representations}, 2025.

\bibitem[Zhang et~al.(2025)Zhang, Hou, Xiao, Huang, and Li]{zhang2025emmu}
Jie Zhang, Jiahui Hou, Tie Xiao, Yunyi Huang, and Xiang-Yang Li.
\newblock Emmu: Efficient information-level multimodal machine unlearning with high model fidelity.
\newblock In \emph{2025 IEEE/ACM 33rd International Symposium on Quality of Service (IWQoS)}, pages 1--10. IEEE, 2025.

\bibitem[Zhang et~al.(2024)Zhang, Lin, Bai, and Mei]{zhang2024negative}
Ruiqi Zhang, Licong Lin, Yu Bai, and Song Mei.
\newblock Negative preference optimization: From catastrophic collapse to effective unlearning.
\newblock In \emph{First Conference on Language Modeling}, 2024.

\end{thebibliography}
}

\end{document}